\documentclass[conference]{IEEEtran}
\IEEEoverridecommandlockouts
\usepackage{cite}
\usepackage{amsmath,amssymb,amsfonts}
\usepackage{bm}
\usepackage{algorithmic}
\usepackage{graphicx}
\usepackage{textcomp}
\usepackage{xcolor}
\usepackage{siunitx}
\usepackage{booktabs}
\usepackage{stfloats}

\def\BibTeX{{\rm B\kern-.05em{\sc i\kern-.025em b}\kern-.08em
    T\kern-.1667em\lower.7ex\hbox{E}\kern-.125emX}}
\begin{document}

\title{PACT: Pose-Robust Actuation Co-Design for Tendon-Driven Robotic Hands
}

\author{
    \IEEEauthorblockN{
        Prashant Kumar$^{\dagger}$,
        Rachael Granberry,
        Taylor Niehues,
        Andrew A. Stanley,
        Gabe Pirie
    }
    \vspace{0.5cm}
    \IEEEauthorblockA{
        Meta Reality Labs Research\\[1em]
        $^{\dagger}$Work done while at Meta.
    }
}

\maketitle

\begin{abstract}
Humanoid robotic hands must provide sufficient force, speed and backdrivability within a human-scale envelope. Tendon-driven transmissions address this challenge by relocating actuators to the palm and forearm, but routing  paths, moment arms, and capstan dimensions jointly determine both force capability and joint speed, so selecting them sequentially satisfies one requirement at the expense of the other. This paper presents a pose-robust actuation co-design for tendon-driven robot hands (PACT) framework that optimizes tendon routing, moment arms, and capstan dimensions under pose-dependent force requirements and drivetrain-bandwidth constraints. A feasible-force formulation enforces actuator bounds and equilibrium at all joints, while a max--min objective maximizes task-force coverage at the weakest pose. Applied to a four-degree-of-freedom finger, the framework produces an asymmetric \(N+1\) tendon topology using quasi-direct-drive actuators. Across eight poses, the design achieves task-force coverage of \(0.88\)--\(1.00\) and a predicted \(5\,\mathrm{Hz}\) minimum isolated-joint flexion/extension bandwidth within a 70th-percentile human-size envelope. Prototype measurements
deviate from predictions by up to \(24.5\%\), primarily due to unmodeled transmission losses. These results indicate that co-designing routing with the drivetrain can meet pose-dependent force requirements without relaxing the bandwidth constraint, paving the way for human-sized robotic hands with human-level performance.
\end{abstract}


\section{Introduction}

Dexterous robotic hands require substantial force capability, rapid response,
and mechanical backdrivability within a compact form factor. Quasi-direct-drive (QDD) actuators
come closest to closing this gap by combining torque density, bandwidth, and
low reflected impedance through relatively low transmission ratios
\cite{QDD, Katz2019}. Human-hand anthropometry severely limits the space available at the finger
joints \cite{Greiner1991,Buchholz1992}. For the actuator considered here, applying
this dimensional constraint limits the estimated output torque to approximately
\(0.09\,\mathrm{N\,m}\). This is substantially below the joint torque estimated
from representative human fingertip-force measurements
\cite{Mathiowetz1985,ValeroCuevas1998}. Sizing a QDD actuator for the maximum load at every joint
exceeds the packaging, mass, and thermal budgets of a human-scale hand. This
tradeoff forces robotic hands either to exceed the dimensions of a human hand or to accept force and bandwidth penalties. Tendon-driven transmissions alleviate
this constraint by relocating actuators proximally, reducing distal mass and
enabling compact fingers without sacrificing useful grasping forces
\cite{Salisbury1982,Jacobsen1984,Grebenstein2011}.

\begin{figure}[hbt!]
    \centering
    \includegraphics[
        width=1\columnwidth,
        trim={0cm 0cm 0cm 0cm},
        clip
    ]{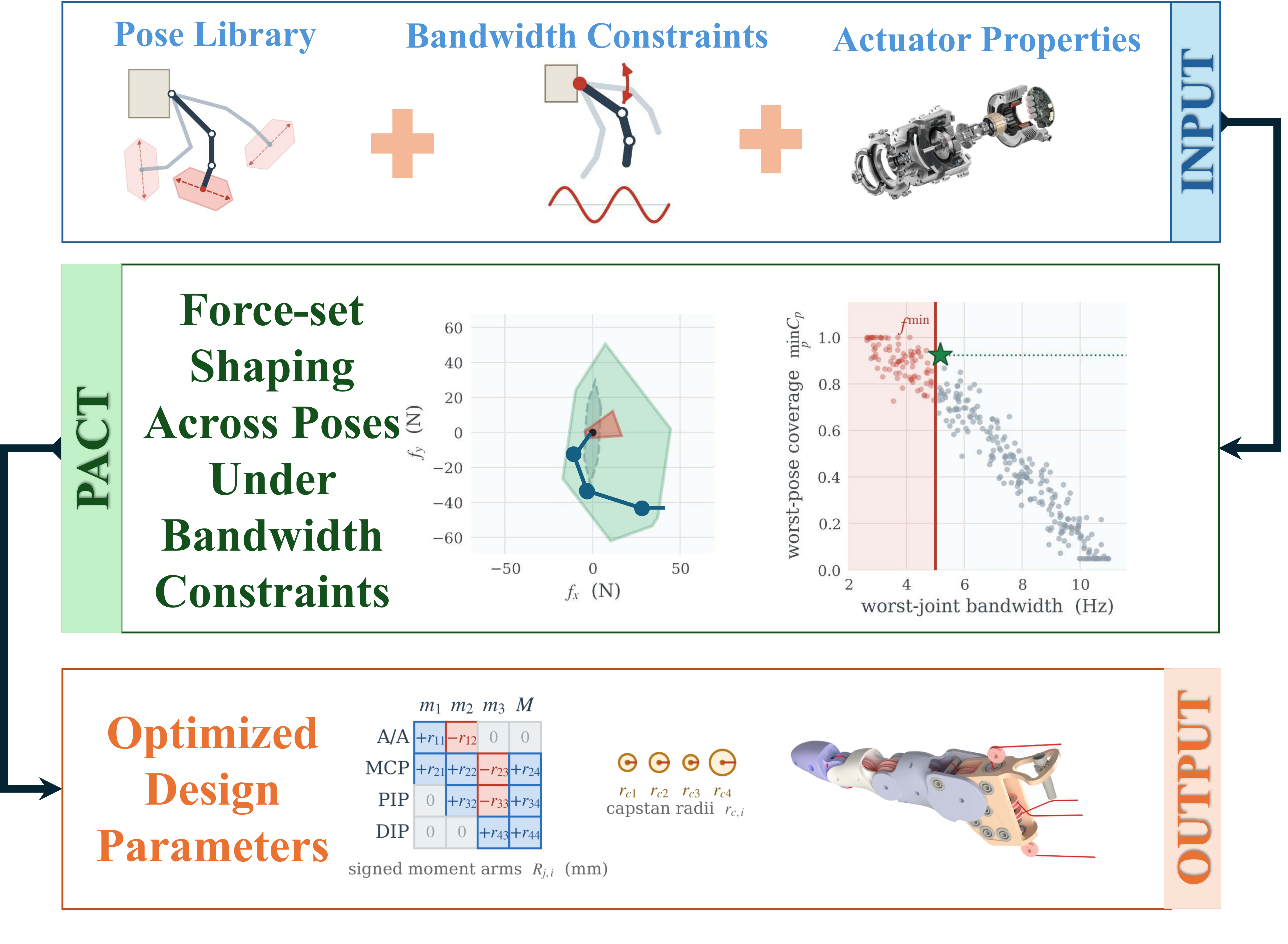}
    \caption{\textbf{Pose-robust actuation co-design for tendon-driven robot hands (PACT) framework.} The framework combines a force-pose library, task-specific bandwidth constraints, and actuator properties to jointly optimize tendon-driven robot hands. It shapes feasible force sets across representative poses, maximizes the workspace coverage under worst-joint bandwidth limits, and selects design parameters that balance force capability, motion speed, and pose robustness. The resulting parameters define the tendon-routing and actuation configuration of the optimized hand.}
    \label{fig:framework}
\end{figure}

The performance of a tendon-driven finger depends on strongly coupled routing
and drivetrain choices. Tendon topology and moment arms map tendon forces to
joint torques, whereas motor selection, gearing, and capstan dimensions
determine tendon-force capacity, motor-speed requirements, and reflected
inertia. For example, reducing the capstan radius increases the available tendon force but proportionally lowers the joint speed obtainable from a given motor speed. Such couplings motivate optimizing the routing and drivetrain parameters jointly, rather than independently, for optimal performance.

While achieving sufficient fingertip force is important, the objective is not necessarily to maximize
the total force-set volume, but to orient and shape the feasible set toward the
required task-force region. Tendon coupling can be exploited to concentrate fingertip-force
capability in task-relevant directions to avoid allocating actuation capacity
to directions that contribute little to the intended manipulation tasks.

Existing force-set optimization frameworks \cite{Inouye2014} generally consider a single pose
and therefore do not guarantee that task requirements are met throughout the
required workspace. Because the manipulator Jacobian varies with joint
configuration, a routing and transmission optimized at one pose may perform
poorly at another. Although prior studies have examined tendon routing,
asymmetric actuation, force-polytope shaping, and actuator distribution
\cite{Pollard2002,Inouye2014,Gosselin1991,Russo2023}, the joint optimization
of tendon topology, heterogeneous actuator allocation, transmission geometry,
force capability, and bandwidth across multiple poses remains insufficiently
explored.

This paper presents a pose-robust actuation co-design for tendon-driven robot hands (PACT) framework that selects tendon
routing and drivetrain parameters simultaneously. Given a library of poses
with associated force and motion requirements, the framework optimizes tendon
topology, signed moment arms, capstan radii, and actuator allocation.
Fingertip-force capability is evaluated using an exact feasible-force set that
enforces all joint-equilibrium equations. Dynamic feasibility is assessed
using a drivetrain model incorporating continuous motor torque, speed limits,
gear ratios, capstan radii, and reflected rotor inertia. A max--min objective
selects the shared physical design that maximizes performance at its weakest
pose.

We apply the framework as shown in Figure~\ref{fig:framework} to a four-degree-of-freedom (DoF) finger with three
flexion/extension (F/E) joints and one MCP abduction/adduction (A/A) DoF.
The resulting design uses an asymmetric \(N+1\) tendon arrangement. A
wrist-mounted, high-torque QDD actuator drives a shared, force-dominant flexion
tendon, while three smaller palm-mounted QDD actuators drive differential
tendons that regulate posture and shape the force set. A separate
bidirectional actuator controls abduction/adduction. This heterogeneous
allocation avoids uniformly oversizing the actuators and limits the number of
high-force tendons crossing the wrist.

In summary, this paper contributes:

\begin{enumerate}
    \item \textbf{Integrated routing and drivetrain design.} A mixed
    discrete--continuous optimization selects routing signs, moment arm
    magnitudes and capstan radii.

    \item \textbf{Worst-pose performance optimization.} A max--min objective
    selects one shared physical design by explicitly maximizing performance at
    the weakest pose in a prescribed pose library.

    \item \textbf{Exact feasible-force evaluation.} Admissible fingertip
    forces are defined as the preimage of the bounded actuator-generated
    joint-torque set under the transposed fingertip Jacobian. Unlike
    pseudoinverse projections, this formulation admits only forces that
    satisfy every joint-equilibrium equation.

    \item \textbf{Experimental validation.} A four-DoF
    tendon-driven finger demonstrates the proposed framework experimentally.
\end{enumerate}

\section{Related Work}

\subsection{Tendon-Driven Transmission and Actuator Placement}

Tendon-driven mechanisms relocate actuators away from the joints, reducing
distal mass and enabling compact robotic fingers
\cite{Salisbury1982,Jacobsen1984,Grebenstein2011}. These benefits come at the
cost of tendon friction, compliance, backlash, pretension requirements, and
routing congestion. Routing several tendons through the wrist can further
increase mechanical complexity and introduce parasitic coupling during wrist
motion.

Distributed architectures place some actuators in the palm while retaining
larger units in the wrist or forearm, shortening tendon paths and reducing
wrist congestion \cite{arXiv2512.04399}. Existing architectures generally
prescribe actuator placement and routing rather than selecting them from task
requirements.

\subsection{Tendon Topology and Heterogeneous Actuator Sizing}

Conventional tendon-driven fingers often use antagonistic tendon pairs,
yielding a \(2N\) architecture for \(N\) independently actuated joints. This
arrangement allocates similar capacity to opposing directions even when task
loads are strongly directional.

Pollard and Gilbert \cite{Pollard2002} showed that an asymmetric \(N+1\)
arrangement can approximate the flexion-biased force capability of the human
finger with fewer tendons than a fully antagonistic design. Related work
assigns actuator capacity according to local load requirements rather than
sizing every actuator identically \cite{arXiv2512.04399}.

Morrell and Salisbury \cite{Morrell1995} proposed a micro--macro architecture
in which large, low-bandwidth and small, high-bandwidth actuators operate in
parallel. In contrast, the present architecture uses high-bandwidth,
backdrivable QDD actuators to drive different tendons within an \(N+1\) topology.
A high-torque actuator supplies the dominant grasping load, while smaller
actuators provide differential joint authority.

\subsection{Force-Set and Mechanism Optimization}

Force polytopes represent the Cartesian forces that a mechanism can generate
under bounded actuator inputs and have been used to assess force capability,
isotropy, and task-direction performance \cite{Gosselin1991}.

Inouye and Valero-Cuevas \cite{Inouye2014} optimized tendon topology and
pulley dimensions for a three-DoF finger using the Maximum
Isotropic Value. Their square, invertible Jacobian allowed joint-torque
polytopes to be mapped directly into Cartesian force space. The finger studied
here instead has four joint coordinates and a three-dimensional Cartesian
force space, so not every achievable joint-torque vector corresponds to a
valid fingertip force. The complete joint-equilibrium equation must therefore
be enforced rather than approximated through pseudoinverse projection.

Previous mechanism-optimization studies have considered routing,
transmission ratios, workspace-level performance, and task-dependent force
metrics \cite{Russo2023}. However, they commonly evaluate a single pose or an
aggregate metric, prescribe part of the actuation architecture, or represent
actuator capability using static torque bounds without accounting for speed
and reflected inertia. The present framework addresses these limitations by
selecting a shared routing and drivetrain design against force and bandwidth
requirements across a prescribed pose library.

\section{Mathematical Model}
\label{sec:models}

\subsection{Finger and Actuation Model}

The finger has four-DoF. The three flexion/extension (MCP, PIP,
and DIP joints) are actuated by an \(N+1\) arrangement of four pull-only
tendons,
\vspace{-0.1cm}
\begin{equation}
\mathbf{T}
=
\begin{bmatrix}
T_{m_1} & T_{m_2} & T_{m_3} & T_M
\end{bmatrix}^{\mathsf T},
\qquad
\mathbf{T}\geq\mathbf{0}.
\end{equation}

\begin{figure}[hbt!]
    \centering
    \includegraphics[
        width=0.8\columnwidth,
        trim={0cm 5cm 2cm 4cm}
    ]{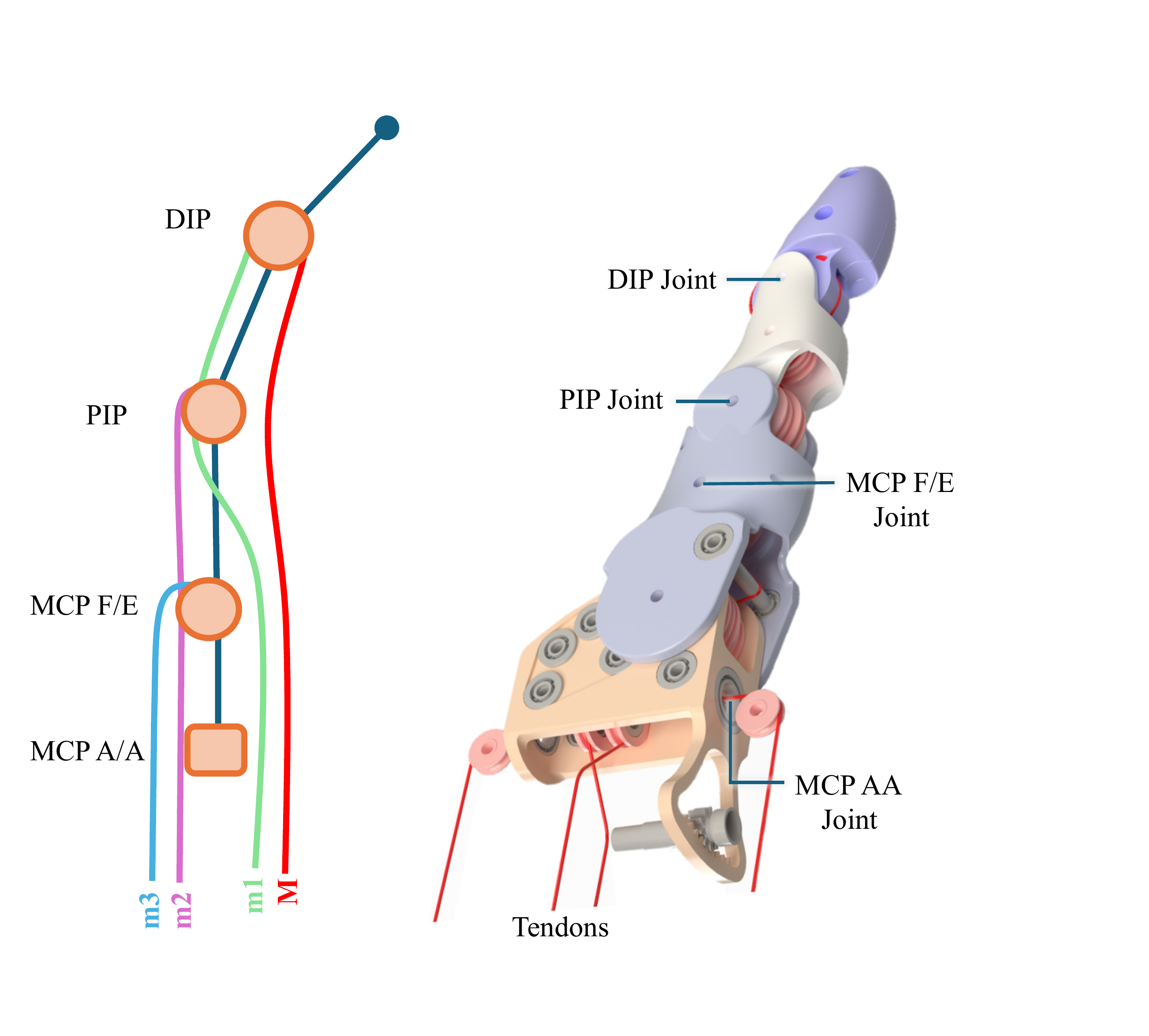}
    \caption{Actuation layout of the four-DoF finger. Flexion/extension uses
    an \(N+1\) tendon arrangement, while MCP abduction/adduction is actuated
    separately.}
    \label{fig:finger_vector}
\end{figure}

Tendons \(m_1\), \(m_2\), and \(m_3\) are driven by low-torque QDD actuators,
whereas \(M\) is driven by a high-torque QDD actuator. A separate
bidirectional actuator supplies the abduction/adduction torque
\(\tau_{\mathrm{AA}}\). Figure~\ref{fig:finger_vector} shows the actuation
layout of the finger architecture.

The signed routing matrix
\(\mathbf{R}\in\mathbb{R}^{4\times4}\) maps tendon tensions to joint torques.
Under a point-contact model, static equilibrium requires
 \vspace{-0.2cm}
\begin{equation}
\mathbf{J}(\mathbf{q})^{\mathsf T}\mathbf{f}
=
\mathbf{R}\mathbf{T}
+\mathbf{e}_{\mathrm{AA}}\tau_{\mathrm{AA}},
\label{eq:equilibrium}
\end{equation}

where \(\mathbf{J}(\mathbf{q})\in\mathbb{R}^{3\times4}\) is the fingertip
Jacobian, \(\mathbf{f}\in\mathbb{R}^{3}\) is the Cartesian fingertip-force
vector, and
\(\mathbf{e}_{\mathrm{AA}}=[1\;0\;0\;0]^{\mathsf T}\). The actuator bounds are
 \vspace{-0.1cm}
\begin{equation}
\mathbf{0}
\leq
\mathbf{T}
\leq
\overline{\mathbf{T}}^{\mathrm{peak}},
\qquad
|\tau_{\mathrm{AA}}|
\leq
\overline{\tau}_{\mathrm{AA}}.
\label{eq:actuatorbounds}
\end{equation}

The peak tension available from tendon \(i\) is
 \vspace{-0.1cm}
\begin{equation}
\overline{T}^{\mathrm{peak}}_i
=
\frac{\tau_{i,\mathrm{peak}}^{\mathrm{out}}\eta_i}{r_{c,i}},
\label{eq:staticlimit}
\end{equation}

where \(\tau_{i,\mathrm{peak}}\), \(\eta_i\), and \(r_{c,i}\) are
the motor peak torque at the gearbox, nominal transmission efficiency, and
capstan radius, respectively.

\subsection{Feasible Force Set}

Equations~(\ref{eq:equilibrium}) and~(\ref{eq:actuatorbounds}) define the
achievable joint-torque set
 \vspace{-0.2cm}
\begin{equation}
\mathcal{T}_{\tau}
=
\left\{
\boldsymbol{\tau}\in\mathbb{R}^{4}
\;\middle|\;
\begin{gathered}
\boldsymbol{\tau}
=\mathbf{R}\mathbf{T}
+\mathbf{e}_{\mathrm{AA}}\tau_{\mathrm{AA}},\\
\mathbf{0}\leq\mathbf{T}\leq\overline{\mathbf{T}}^{\mathrm{peak}},
\quad
|\tau_{\mathrm{AA}}|\leq\overline{\tau}_{\mathrm{AA}}
\end{gathered}
\right\}.
\label{eq:torqueset}
\end{equation}

At pose \(\mathbf{q}\), the exact feasible fingertip-force set is

\begin{equation}
\mathcal{F}(\mathbf{q})
=
\left\{
\mathbf{f}\in\mathbb{R}^{3}
\;\middle|\;
\mathbf{J}(\mathbf{q})^{\mathsf T}\mathbf{f}
\in
\mathcal{T}_{\tau}
\right\}.
\label{eq:strictset}
\end{equation}

Because
\(\mathbf{J}(\mathbf{q})^{\mathsf T}\in\mathbb{R}^{4\times3}\), not every
achievable joint-torque vector corresponds to a Cartesian fingertip force.
Equation~(\ref{eq:strictset}) enforces all equilibrium equations, whereas a
pseudoinverse projection may discard unmatched torque and overestimate force
capability.

Writing the torque set in half-space form,
 \vspace{-0.1cm}
\begin{equation}
\mathcal{T}_{\tau}
=
\left\{
\boldsymbol{\tau}\in\mathbb{R}^{4}
\;\middle|\;
\mathbf{a}_k^{\mathsf T}\boldsymbol{\tau}
\leq h_k,
\quad
k=1,\ldots,N_h
\right\},
\label{eq:torquehalfspaces}
\end{equation}

and substituting
\(\boldsymbol{\tau}
=\mathbf{J}(\mathbf{q})^{\mathsf T}\mathbf{f}\) gives

\begin{equation}
\mathcal{F}(\mathbf{q})
=
\left\{
\mathbf{f}\in\mathbb{R}^{3}
\;\middle|\;
\bigl(\mathbf{J}(\mathbf{q})\mathbf{a}_k\bigr)^{\mathsf T}
\mathbf{f}
\leq h_k,
\quad
k=1,\ldots,N_h
\right\}.
\label{eq:forcehalfspaces}
\end{equation}

Singular or lower-dimensional force sets are treated as invalid rather than
approximated using a pseudoinverse.

For a unit direction \(\mathbf{d}\), the direction-constrained force
capability is
 \vspace{-0.2cm}
\begin{equation}
\begin{aligned}
F_{\mathrm{dc}}(\mathbf{q},\mathbf{d})
={}&
\max_{\alpha,\mathbf{T},\tau_{\mathrm{AA}}}
\quad \alpha\\
\mathrm{s.t.}\quad
&
\mathbf{J}(\mathbf{q})^{\mathsf T}(\alpha\mathbf{d})
=
\mathbf{R}\mathbf{T}
+\mathbf{e}_{\mathrm{AA}}\tau_{\mathrm{AA}},\\
&
\mathbf{0}
\leq
\mathbf{T}
\leq
\overline{\mathbf{T}}^{\mathrm{peak}},\\
&
|\tau_{\mathrm{AA}}|
\leq
\overline{\tau}_{\mathrm{AA}},
\qquad
\alpha\geq0.
\end{aligned}
\label{eq:directional_force}
\end{equation}

Unlike a support function, this formulation constrains the complete force
vector to lie along \(\mathbf{d}\). The resulting linear programs are solved
using HiGHS~\cite{HuangfuHall2018} and verified against the original
equilibrium and actuator constraints.

\subsection{Drivetrain Bandwidth}

Static force capability uses peak actuator-output torque, whereas bandwidth is
evaluated using continuous output torque, output-speed limits, and reflected
motor-rotor inertia. Each joint is analyzed in isolation: joint \(j\) is
commanded along the prescribed trajectory while all remaining joints are held
fixed. This approximation reduces the tendon-coupled dynamics to independent
single-DoF problems. During isolated motion, only tendons producing torque in the commanded direction are assumed taut. Antagonistic tendons are assumed to pay out and are therefore treated as kinematically disengaged; their rotor inertias are omitted.
For joint \(j\),
 \vspace{-0.2cm}
\begin{equation}
q_j(t)=q_{j,0}+A_j\sin(\omega_e t),
\end{equation}
where \(A_j\) and \(\omega_e\) are the motion amplitude and excitation
frequency. Let
\(g_i=\omega_{m,i}/\omega_i^{\mathrm{out}}\) denote the gearbox reduction
ratio. The joint-to-output and joint-to-rotor speed ratios are
\begin{equation}
G_{j,i}^{\mathrm{out}}
=
\frac{|R_{j,i}|}{r_{c,i}},
\qquad
G_{j,i}^{\mathrm{rot}}
=
g_iG_{j,i}^{\mathrm{out}},
\end{equation}
and the peak gearbox-output speed is
\begin{equation}
\omega_{j,i}^{\mathrm{out,peak}}
=
G_{j,i}^{\mathrm{out}}A_j\omega_e.
\label{eq:motorspeed}
\end{equation}
For motion direction \(\sigma\in\{-1,+1\}\), the active tendons are
\begin{equation}
\mathcal{A}_{j,\sigma}
=
\{i\mid \sigma R_{j,i}>0\}.
\end{equation}
The effective inertia and required peak torque are
\begin{align}
I_{j,\sigma}^{\mathrm{eff}}
&=
I_j^{\mathrm{load}}
+
\sum_{i\in\mathcal{A}_{j,\sigma}}
J_{m,i}\left(G_{j,i}^{\mathrm{rot}}\right)^2,
\\
\tau_{j,\sigma}^{\mathrm{req}}(\omega_e)
&=
I_{j,\sigma}^{\mathrm{eff}}A_j\omega_e^2,
\label{eq:isolatedbw}
\end{align}
where \(J_{m,i}\) is the motor rotor inertia. The continuous tendon-force
limit is
\begin{equation}
\overline{T}_{j,i}^{\mathrm{cont}}(\omega_e)
=
\frac{
\eta_i\overline{\tau}_i^{\mathrm{out,cont}}
\left(\omega_{j,i}^{\mathrm{out,peak}}\right)
}{
r_{c,i}
},
\end{equation}
and the available joint torque is
\begin{equation}
\tau_{j,\sigma}^{\mathrm{avail}}(\omega_e)
=
\sum_{i\in\mathcal{A}_{j,\sigma}}
|R_{j,i}|\,
\overline{T}_{j,i}^{\mathrm{cont}}(\omega_e).
\end{equation}
The bandwidth is the highest frequency satisfying
\(\tau_{j,\sigma}^{\mathrm{req}}
\leq\tau_{j,\sigma}^{\mathrm{avail}}\) and
\(\omega_{j,i}^{\mathrm{out,peak}}
\leq\overline{\omega}_i^{\mathrm{out}}\) for all active tendons. It is found
by bracketing and bisection, with
\begin{equation}
f_{j,\sigma}^{\mathrm{bw}}
=
\frac{\omega_{j,\sigma}^{\mathrm{bw}}}{2\pi}.
\end{equation}
Flexion and extension are evaluated separately for the MCP, PIP, and DIP
joints.

\section{Optimization Framework}
\label{sec:optimization}

The framework maximizes task-force coverage over poses
\(\mathcal{P}=\{1,\ldots,N_p\}\). At pose \(p\), the feasible force polytope is
\(\mathcal{F}(\mathbf{q}_p)\), and the required force region is
 \vspace{-0.2cm}
\[
\mathcal{R}_p=
\operatorname{conv}\left\{
\rho_p(\mathbf{d}_k)\mathbf{d}_k
\right\},
\]
where \(\rho_p(\mathbf{d}_k)\) is the required magnitude in direction
\(\mathbf{d}_k\). Coverage is
\begin{equation}
C_p(\mathbf{x})=
\frac{
\mu\!\left(\mathcal{R}_p\cap\mathcal{F}(\mathbf{q}_p)\right)
}{
\mu\!\left(\mathcal{R}_p\right)
},
\label{eq:coverage}
\end{equation}
where \(\mu\) denotes force-space volume. Thus, \(C_p=1\) indicates full
coverage, while directional task asymmetry is captured directly by
\(\mathcal{R}_p\).

\subsection{Design variables}
Task-force coverage is raised by changing the transmission, not the actuators. Tendon routing sets which joints each tendon drives and in which direction. Moment arm magnitudes set how strongly tendon force converts to joint torque. Capstan radii set the tendon force available from a given motor torque. These three quantities reshape and reorient \(\mathcal{F}(\mathbf{q}_p)\), and the framework selects them together. Actuator assignments are held fixed within each candidate design and enumerated alongside the routing topologies. The design vector is

\begin{equation}
\mathbf{x}=
\begin{bmatrix}
\operatorname{vec}(\mathbf{R})^{\mathsf T} &
r_{c,1} & r_{c,2} & r_{c,3} & r_{c,4}
\end{bmatrix}^{\mathsf T}.
\label{eq:designvector}
\end{equation}

where \(\operatorname{vec}(\cdot)\) stacks the free entries of \(\mathbf{R}\) column-wise. Each free routing entry is parameterized as

\begin{equation}
R_{j,i}=s_{j,i}r_{j,i},
\qquad
s_{j,i}\in\{-1,0,+1\},
\label{eq:routingdesign}
\end{equation}

where \(s_{j,i}\) selects the torque direction or removes the route, and
\(r_{j,i}\) is the moment arm magnitude. Entries fixed by the mechanical
architecture, including the prescribed abduction/adduction couplings, are not
varied. The continuous variables satisfy

\begin{equation}
r_{j,i}^{\min}\leq r_{j,i}\leq r_{j,i}^{\max},
\qquad
r_{c,i}^{\min}\leq r_{c,i}\leq r_{c,i}^{\max}.
\label{eq:designbounds}
\end{equation}

Bounds are set by the finger geometry and available size envelope.

\subsection{Pose Objective}

At pose \(p\), the design score is the task-force coverage
\begin{equation}
S_p(\mathbf{x}) = C_p(\mathbf{x}),
\label{eq:poseobjective}
\end{equation}
where \(C_p(\mathbf{x})\), defined in
(\ref{eq:coverage}), is the fraction of the required force region contained
within the feasible fingertip-force set. Thus, \(S_p=1\) indicates full
coverage of the pose-specific force requirement.

The design must also satisfy
\begin{equation}
f_{j,\sigma}^{\mathrm{bw}}(\mathbf{x})
\geq f_{j,\sigma}^{\min},
\quad
j\in\{\mathrm{MCP},\mathrm{PIP},\mathrm{DIP}\},
\quad
\sigma\in\{+,-\},
\label{eq:bandwidthconstraints}
\end{equation}
where \(\sigma\) denotes flexion or extension and
\(f_{j,\sigma}^{\min}\) is the required minimum isolated-joint bandwidth.
Abduction/adduction and coordinated-motion bandwidths are retained as
diagnostic quantities.

\subsection{Worst-Pose Objective}

Because one physical design must operate at every pose, the optimization
maximizes the minimum task-force coverage:
\begin{align}
\max_{\mathbf{x},z}\quad
& z
\label{eq:sharedoptimization}\\
\mathrm{s.t.}\quad
& z \leq C_p(\mathbf{x}),
&& \forall p\in\mathcal{P}, \notag\\
& \mathcal{F}(\mathbf{q}_p)\ \text{is valid},
&& \forall p\in\mathcal{P}, \notag\\
& f_{j,\sigma}^{\mathrm{bw}}(\mathbf{x})
  \geq f_{j,\sigma}^{\min},
&& \forall j,\sigma, \notag\\
& s_{j,i}\in\{-1,0,+1\},
&& \forall j,i, \notag\\
& \mathbf{x}\ \text{satisfies (\ref{eq:designbounds})}. &&
\notag
\end{align}

At the optimum,
\begin{equation}
z^\star = \min_{p\in\mathcal{P}} C_p(\mathbf{x}^\star),
\end{equation}
so \(z^\star\) is the coverage achieved at the weakest pose. Maximizing this
quantity prevents strong performance at favorable poses from masking poor
performance elsewhere.

\subsection{Solution Procedure}
Admissible routing topologies and actuator assignments are first enumerated. For each topology,
deterministic anchor points and scrambled Sobol samples
\cite{Sobol1967} generate candidate moment arms and capstan radii. Candidates
are rejected if any pose produces an invalid force set or if any bandwidth
constraint is violated. The feasible candidates are ranked by

\begin{equation}
z(\mathbf{x})=\min_{p\in\mathcal{P}}S_p(\mathbf{x}).
\label{eq:worstposescore}
\end{equation}

The best candidates from each topology are refined using bounded adaptive
Nelder--Mead optimization \cite{GaoHan2012}. Finalists are then evaluated at
full numerical resolution and checked against the equilibrium and actuator
bounds. The procedure is deterministic for a fixed configuration but remains
heuristic and does not certify global optimality.

\section{Case Study}
\label{sec:optimization_study}

The case study evaluates the optimization framework on a representative task library to characterize the solution space it produces. The exercise assesses how much a shared design improves on a task-agnostic baseline and how much coverage is given up relative to designs optimized for individual poses. 

Commercial robotic-hand specifications often emphasize scalar performance
measures such as payload or fingertip force rather than task-space force
coverage. Following this convention, the task-agnostic baseline was designed
to produce at least \(15\,\mathrm{N}\) of peak normal fingertip force in the
fully extended pose, without optimizing a force-polytope or task-coverage
metric. Its routing signs were selected
manually to provide flexion and extension authority at each joint, while its
moment arms and capstan radii were adjusted to satisfy all six isolated-joint bandwidth requirements.

\subsection{Task Inputs}
\label{sec:tasks}

The framework accepts pose-specific force directions, limits,
weights, and motion-speed requirements. Eight representative poses span pinch, tripod, extended, deeply flexed, and abducted configurations (Table~\ref{tab:poses}).

\begin{table}[hbt!]
\centering
\caption{Index-finger task poses in degrees.}
\label{tab:poses}
\footnotesize
\begin{tabular}{lcccc}
\toprule
Pose & MCP A/A & MCP F/E & PIP & DIP \\
\midrule
Pose 1: Small tip pinch  & 0  & 60 & 60 & 30 \\
Pose 2: Medium pad pinch & 0  & 45 & 45 & 20 \\
Pose 3: Large pad pinch  & 0  & 25 & 30 & 15 \\
Pose 4: Medium tripod    & 0  & 40 & 50 & 25 \\
Pose 5: Small tripod     & 0  & 55 & 60 & 30 \\
Pose 6: Extended reach   & 0  & 15 & 20 & 10 \\
Pose 7: Deep flexion     & 0  & 70 & 70 & 35 \\
Pose 8: Abducted pinch   & 15 & 45 & 45 & 20 \\
\bottomrule
\end{tabular}
\end{table}

At each pose, the required force region comprises opposing truncated Coulomb
friction cones \cite{Murray1994} aligned with the distal-link palmar normal,
with \(\mu_p=0.5\) and palmar and dorsal limits of \(15\,\mathrm{N}\) and
\(4\,\mathrm{N}\), respectively. A minimum \(5\,\mathrm{Hz}\) isolated-joint
bandwidth is imposed for flexion and extension of the MCP, PIP, and DIP joints.

\subsection{Design Parameters}

The motors for the architecture were fixed in advance as were the link dimensions and task definitions. The optimization framework determines the remaining parameters: routing signs, moment arm magnitudes, and capstan radii,  bounded by a 70th-percentile human-finger
envelope. Table~\ref{tab:actuator_parameters} lists the actuator properties. The modeled finger has a mass of \(0.08\,\mathrm{kg}\), an
abduction/adduction link length of \(a=\SI{10}{\milli\metre}\), and MCP, PIP,
and DIP link lengths of
\(\boldsymbol{\ell}=(45,25,20)\,\si{\milli\metre}\), respectively.

\begin{table}[hbt!]
\centering
\caption{Actuator parameters used in the optimization. All torque and speed
values are specified at the gearbox output.}
\label{tab:actuator_parameters}
\footnotesize
\setlength{\tabcolsep}{4pt}
\begin{tabular}{lcc}
\toprule
Parameter & Low-torque QDD & High-torque QDD \\
\midrule
Motor & GB2204 & Robstride QDD-05 \cite{RobStride05} \\
Gear ratio & \(5{:}1\) & \(7.75{:}1\) \\
Motor Rotor inertia (\(\mathrm{kg\,m^2}\)) &
\(6.8\times10^{-7}\) & \(3.0\times10^{-6}\) \\
Continuous  torque (\(\mathrm{N\,m}\)) & 0.14 & 1.14 \\
Peak  torque (\(\mathrm{N\,m}\)) & 0.40 & 5.4 \\
Output speed (\(\mathrm{rpm}\)) & 512 & 480 \\
Assigned tendons & 3 & 1 \\
\bottomrule
\end{tabular}
\end{table}
\begin{figure}[b!]
    \centering
    \includegraphics[
        width=0.85\columnwidth,
        trim={0cm 0cm 0cm 0.1cm},
        clip
    ]{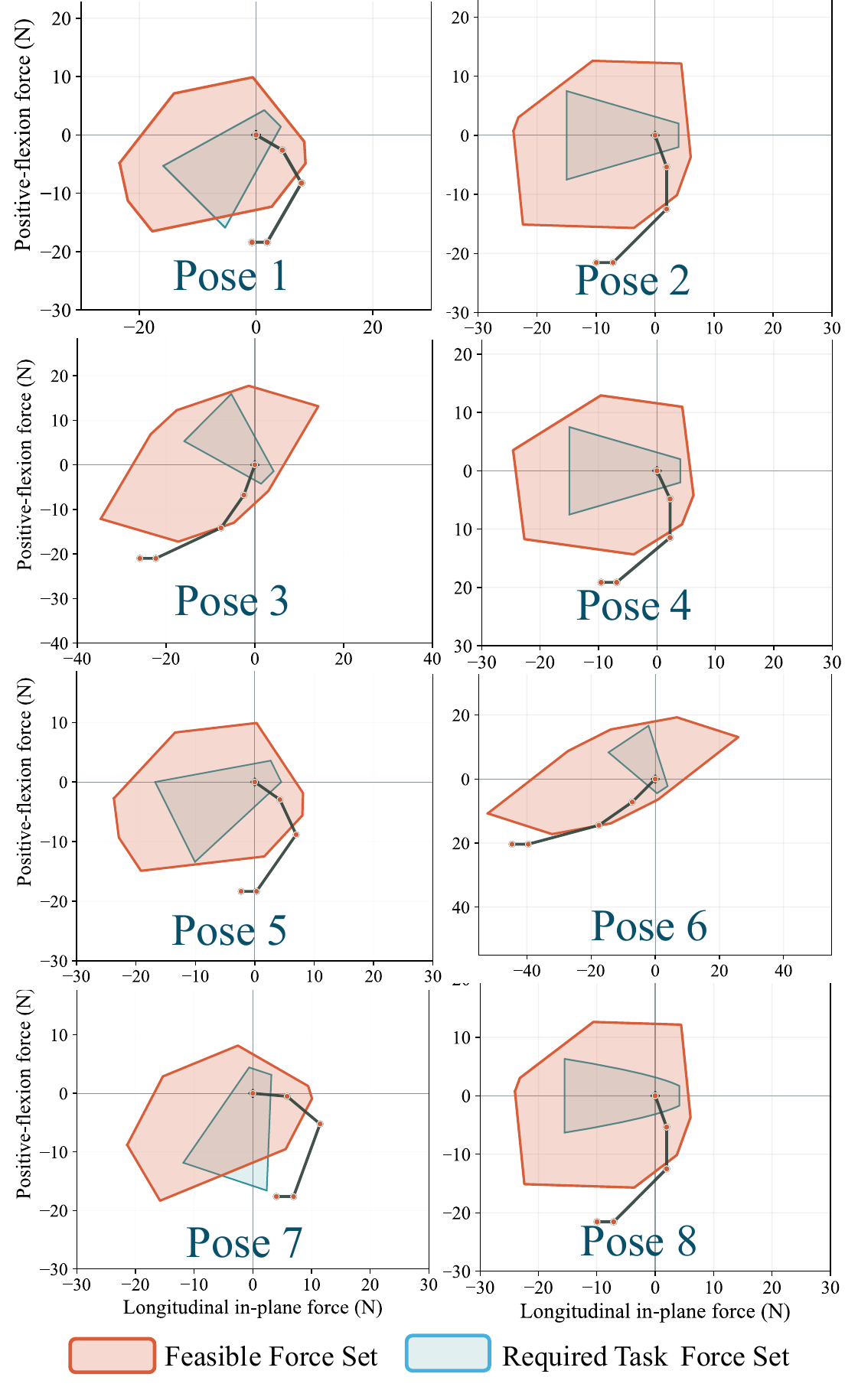}
    \caption{Flexion/extension-plane cross-sections of the feasible force polytopes for the optimized shared design across all eight poses.}
    \label{fig:force_set_across poses}
\end{figure}

\subsection{Optimization Results}
\label{sec:optimization-results}

Figure \ref{fig:force_set_across poses} shows the flexion/extension-plane cross-sections of the feasible force polytopes for the optimized shared design across all eight poses. The shared design raises worst-pose task-force coverage from 0.24 for the baseline to 0.88 and improves every pose in the library (Table~\ref{tab:primaryresults}). At its weakest pose it falls 0.12 below the design optimized for that pose alone, while satisfying the \(5\,\mathrm{Hz}\) bandwidth requirement at all eight poses. All independently optimized poses were able to achieve full coverage.

\begin{table}[hbt!]
\centering
\caption{Results across the eight-pose task library.}
\label{tab:primaryresults}
\footnotesize
\setlength{\tabcolsep}{3pt}
\begin{tabular}{lccc}
\toprule
Quantity & Baseline & Independent & Shared \\
\midrule
Minimum reported score & 0.24 & 1.00 & 0.88 \\
Maximum reported score & 0.69 & 1.00 & 1.00 \\
Minimum F/E bandwidth (Hz) & 5.0 & 5.0 & 5.0 \\
\bottomrule
\end{tabular}
\end{table}

\begin{figure}[hbt!]
    \centering
    \includegraphics[
        width=\columnwidth,
        trim={0cm 11cm 0cm 0.1cm},
        clip
    ]{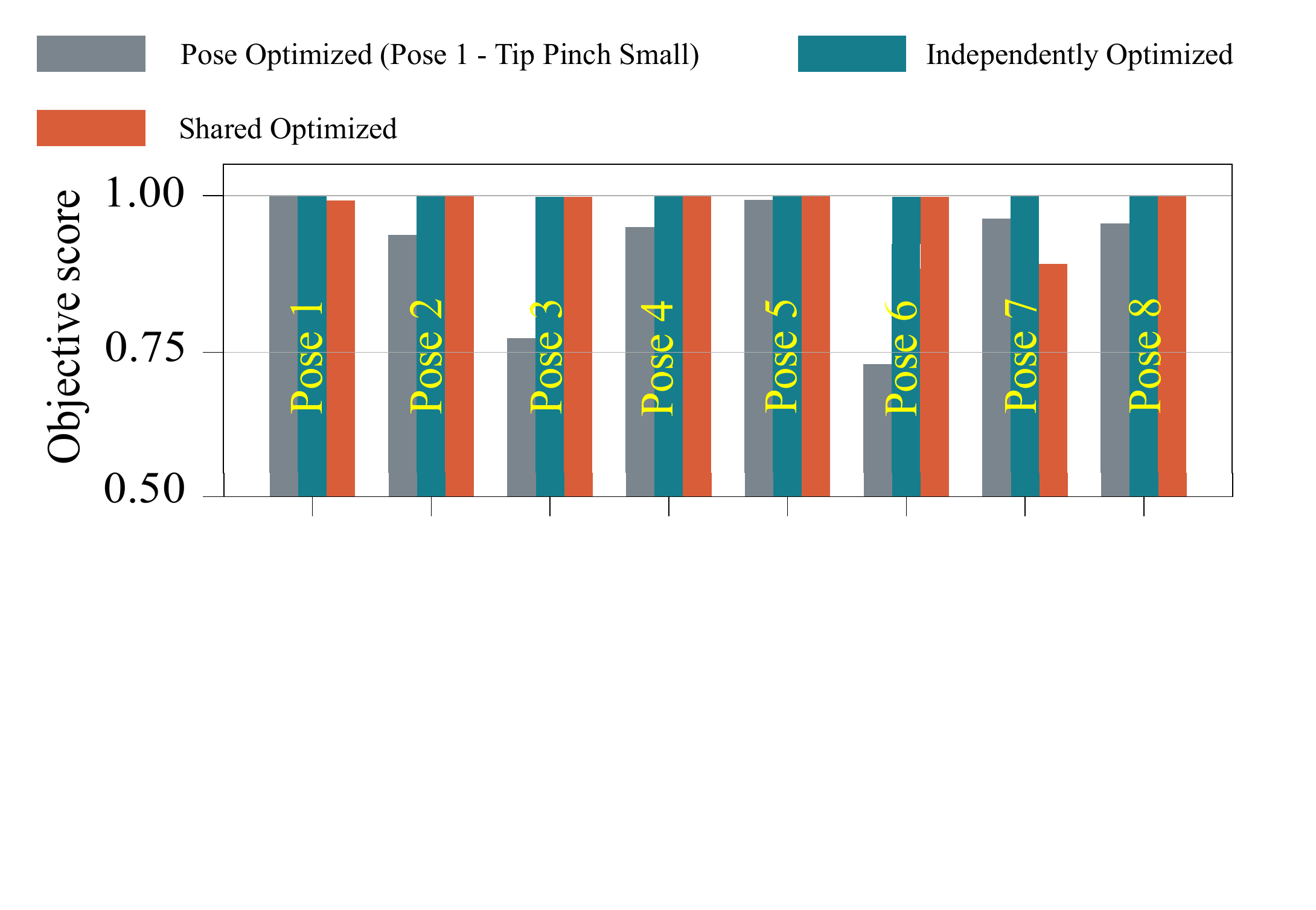}
    \caption{Cross-pose performance of the Pose~1-specific design compared
    with the shared design.}
    \label{fig:cross_pose_scores}
\end{figure}

\begin{figure}[hbt!]
    \centering
    \includegraphics[
        page=6,
        width=\columnwidth,
        trim={0cm 11cm 0cm 0.1cm},
        clip
    ]{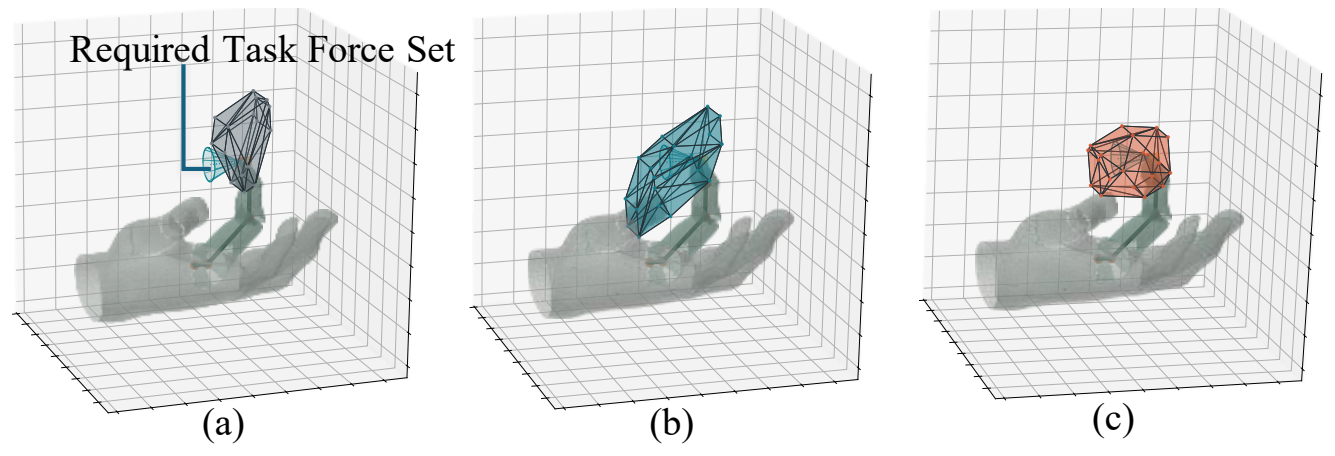}
    \caption{Force sets at Pose~1: (a) baseline, (b) pose-specific design,
    and (c) shared design.}
    \label{fig:force_set_pose2}
\end{figure}
A design optimized for one pose need not perform well elsewhere.
Figure~\ref{fig:cross_pose_scores} shows that the design optimized for Pose 1 (shown in grey) loses
coverage at configurations with different Jacobians, whereas the shared design
performs uniformly across the library. Similar behavior was observed across all the independently optimized designs. Figure~\ref{fig:force_set_pose2} compares the three 3D force sets at Pose~1. The pose-specific design is the most specialized, while the shared design achieves comparable coverage there and also satisfies the other seven poses. The optimized shared routing matrix and capstan diameters are given in Table~\ref{tab:shared_design}. A positive sign indicates that the tendon actuates the joint in flexion, whereas a negative sign indicates extension. The shared design differs from the independently optimized designs because the mapping from fingertip force to joint torque varies with posture.

\begin{table}[hbt!]
\centering
\caption{Optimized shared-design parameters.}
\label{tab:shared_design}
\footnotesize
\setlength{\tabcolsep}{4pt}
\begin{tabular}{lrrrr}
\toprule
Joint Moment Arm(mm) & \(m_1\) & \(m_2\) & \(m_3\) & \(M\) \\
\midrule
MCP A/A (fixed during Optimization ) & \(+3.0\) & \(-3.0\) & \(0.0\) & \(0.0\) \\
MCP F/E  & \(+6.9\) & \(-3.0\) & \(-8.0\) & \(+5.6\) \\
PIP F/E  & \(-3.4\) & \(-3.0\) & \(0.0\) & \(+4.6\) \\
DIP F/E & \(-6.9\) & \(0.0\) & \(0.0\) & \(+3.1\) \\
\midrule
Capstan diameter & \(12.7\) & \(11.0\) & \(13.0\) & \(36.0\) \\
\bottomrule
\end{tabular}
\end{table}

The Deep flexion pose (Pose 7) remains the limiting pose, its kinematics providing reduced mechanical leverage, and MCP extension binds the bandwidth constraints at \({5.0}\, \mathrm{Hz}\).


\section{Experimental Validation}
\label{sec:hardware}

\begin{figure}[hbt!]
    \centering
    \includegraphics[
        width=0.7\columnwidth,
        trim={0cm 3.5cm 0cm 0cm},
        clip
    ]{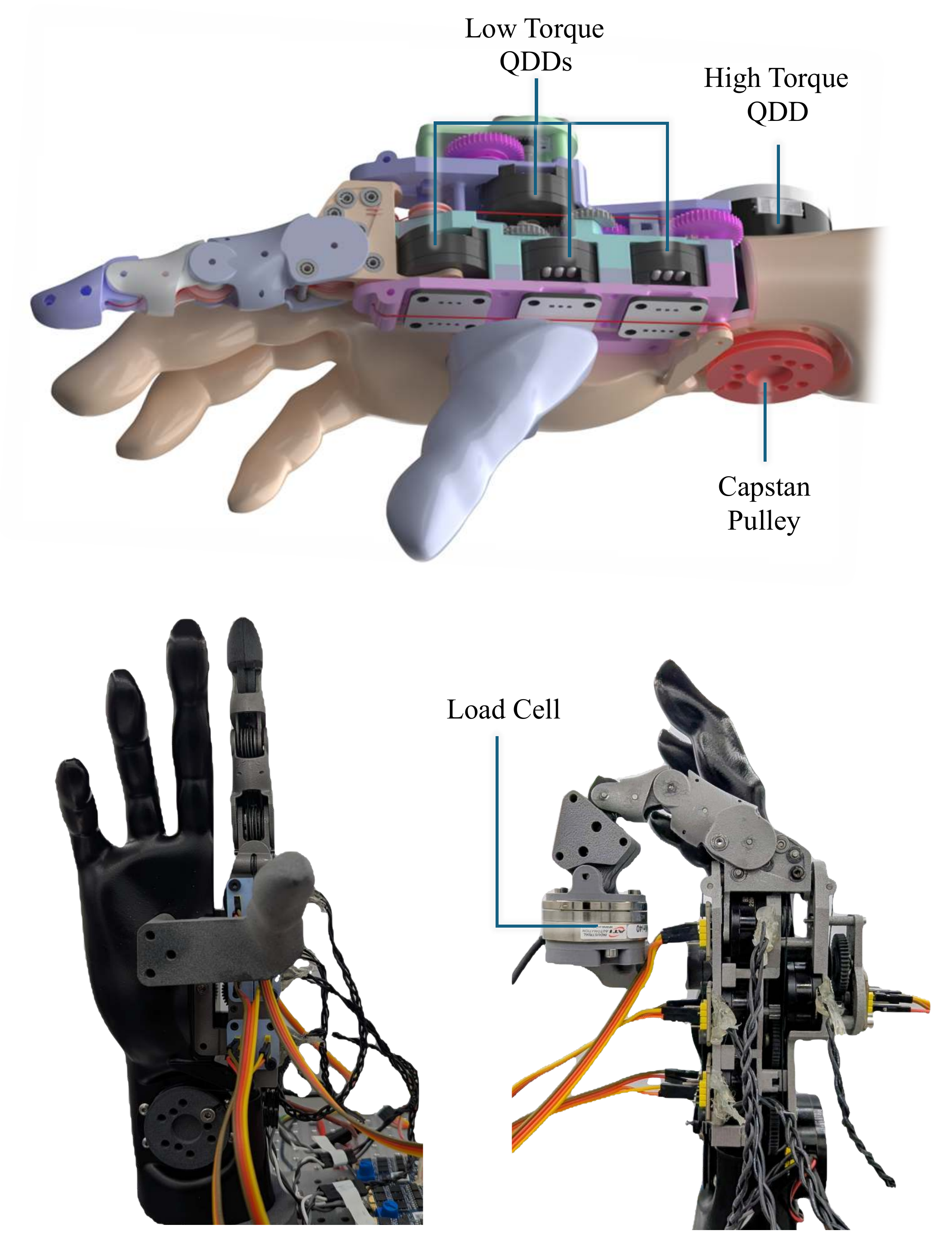}
    \caption{Prototype and experimental setup with fixed finger base and six-axis fingertip load cell}
    \label{fig:4dof_finger_CAD}
\end{figure}

The optimized shared design was realized as the four-DoF prototype shown in
Figure~\ref{fig:4dof_finger_CAD}. The experiments evaluate the predictions of the analytical model regarding the feasible fingertip-force set.

\subsection{Prototype Implementation}

The prototype implements the shared routing matrix and capstan dimensions
shown in Table~\ref{tab:shared_design}. The physical moment arms of the
flexion tendons follow the optimized signs and magnitudes. The
abduction/adduction couplings, however, are fixed by the mechanical geometry
and were not included as free optimization variables.

Four QDD actuators drive the \(N+1\) flexion tendon arrangement. One
high-torque actuator drives tendon \(M\), which crosses all three flexion
joints and supplies most of the grasping torque. Three low-torque actuators
drive tendons \(m_1\), \(m_2\), and \(m_3\), providing the differential
authority required to regulate posture and shape the fingertip-force set. A
fifth bidirectional actuator, with a \(10{:}1\) reduction, drives MCP
abduction/adduction.

Tendons \(m_3\) and \(M\) as shown in Figure~\ref{fig:finger_vector} run approximately parallel to the
abduction/adduction axis and therefore have negligible moment arms.
Tendons \(m_1\) and \(m_2\) pass symmetrically on opposite sides of the axis,
producing small couplings with opposite signs. This arrangement minimizes
lateral torque coupling during flexion and leaves abduction/adduction
predominantly under the authority of its dedicated actuator.

\subsection{Experimental Setup}

Figure~\ref{fig:4dof_finger_CAD} shows the experimental setup. The finger base
is fixed, and a six-axis load cell (Mini40, ATI Industrial Automation) measures
fingertip force through a low-friction pin joint that approximates planar point
contact. Measurements are transformed into the finger frame. Joint angles are
estimated from encoder measurements using calibrated tendon kinematics. Tendon
tension is estimated as
\begin{equation}
\widehat{T}_i=\frac{\kappa_i I_i}{r_{c,i}},
\label{eq:experimental_tension}
\end{equation}
where \(\kappa_i\) is the calibrated gearbox-output torque constant. During each force trial, the tested tendons are commanded in current control
up to their prescribed limits, while the remaining tendons are operated in
position control to maintain the target posture. The measured currents of all
tendons, including those holding the posture, are included in the predicted
joint torques.

\subsection{Fingertip-Force Validation}

The force-set experiment is conducted at
\begin{equation}
\mathbf{q}_{\mathrm{test}}
=
\begin{bmatrix}
0^\circ & 60^\circ & 60^\circ & 30^\circ
\end{bmatrix}^{\mathsf T},
\label{eq:force_test_pose}
\end{equation}
ordered as MCP A/A, MCP F/E, PIP F/E, and DIP F/E. For the test, the finger is moved to the prescribed posture. The
selected tendon commands are then increased gradually to their current limits while other tendons are still holding the pose, and the steady-state fingertip force is recorded by the load cell. Each measurement is averaged over a steady-state interval of
\(5\,\mathrm{s}\) and repeated 5 times per actuator combination.

\begin{figure}[hbt!]
    \centering
    \includegraphics[
        width=0.65\columnwidth,
        trim={0cm 0cm 0cm 0cm},
        clip
    ]{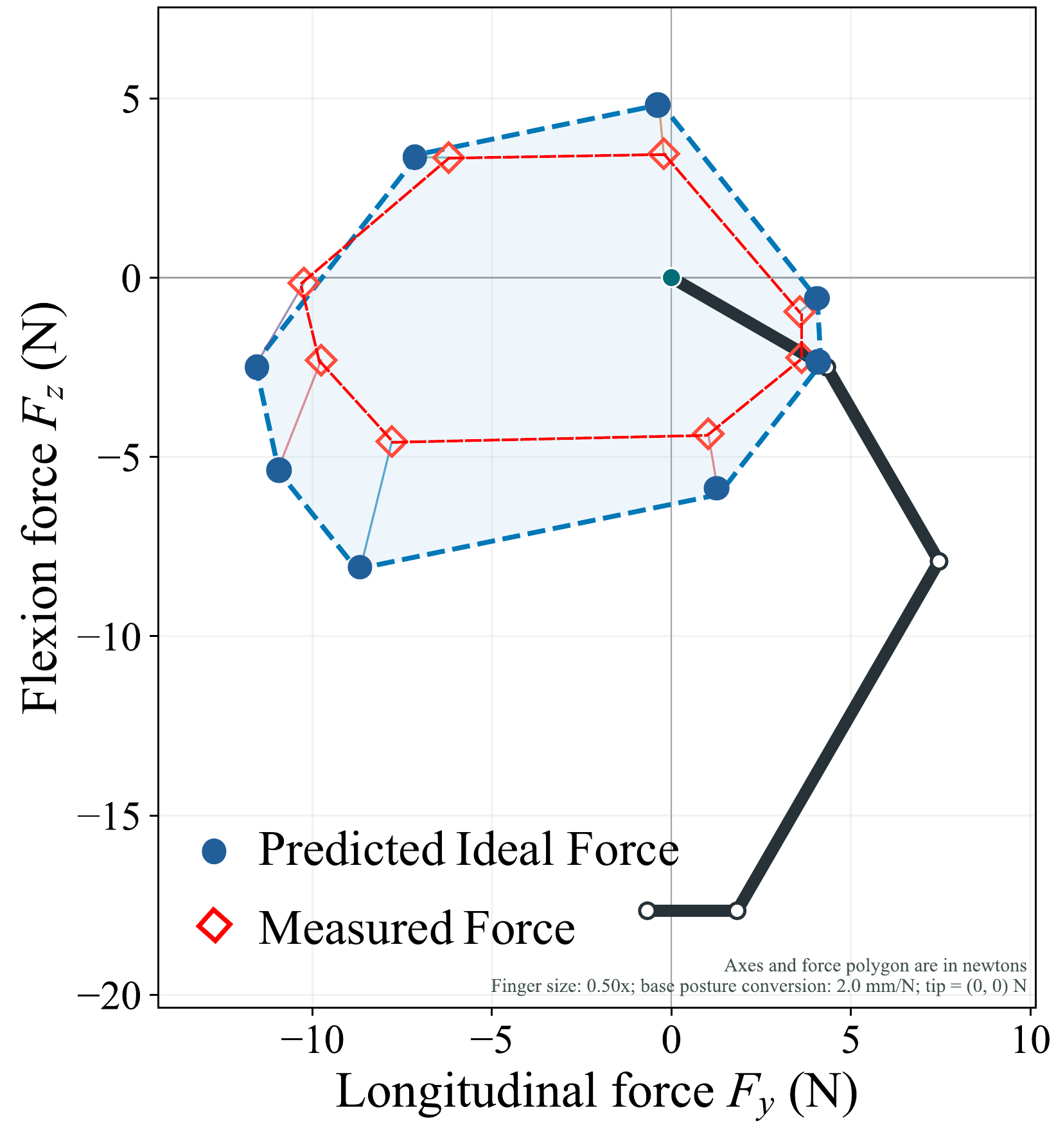}
    \caption{Predicted and measured fingertip forces at
    \(\mathbf{q}_{\mathrm{test}}\). The blue polytope represents the predicted
    force set at the experimental actuator limits. Circles indicate predicted
    forces, and diamonds indicate measurements.}
    \label{fig:force}
\end{figure}

To protect the actuators, the experimental current limits are set to \(50\%\)
of peak torque. The tested actuator combinations correspond to selected vertices of the
feasible force set at these reduced limits. Figure~\ref{fig:force} compares the predicted and measured fingertip forces.

\vspace{-0.3cm}

\subsection{Effect of Tendon-Transmission Losses}

The analytical model assumes that the actuator-side tendon tension is
transmitted uniformly to every crossed joint. In the prototype, sliding
friction, pulley-bearing resistance, guide contact, misalignment, and tendon
compliance reduce the transmitted tension. These effects are represented by
an effective joint-dependent transmission efficiency,
\begin{equation}
T_{i,j}=\eta_{i,j}T_{i,0},
\qquad 0<\eta_{i,j}\leq1,
\label{eq:effective_tendon_efficiency}
\end{equation}
where \(T_{i,0}\) is the actuator-side tension and \(\eta_{i,j}\) is the
cumulative efficiency of tendon \(i\) before reaching joint \(j\). The
delivered joint torque is therefore
\begin{equation}
\tau_j=\sum_i R_{j,i}\eta_{i,j}T_{i,0}.
\label{eq:transmission_loss_torque}
\end{equation}

The measured fingertip forces were mapped to joint torques using
\begin{equation}
\boldsymbol{\tau}^{\mathrm{meas}}_k
=
\mathbf{J}(\mathbf{q}_{\mathrm{test}})^{\mathsf T}
\mathbf{f}^{\mathrm{meas}}_k
\end{equation}
and compared with the nominal frictionless prediction
\(\boldsymbol{\tau}^{\mathrm{pred}}_k=\mathbf{R}\mathbf{T}_k\).
The estimated cumulative torque losses were approximately \(14\%\),
\(24\%\), and \(32\%\) at the MCP, PIP, and DIP joints, respectively,
corresponding to effective transmission efficiencies of \(0.86\), \(0.76\),
and \(0.68\).

The increasing loss toward the distal joints is consistent with the tendon
routing: tendons reaching the PIP and DIP joints encounter longer paths and
additional pulley and guide contacts. Because the losses differ among joints
and tendon paths, the realized joint-torque vector is not a uniformly scaled
version of the frictionless prediction, producing errors in both force magnitude and direction. The measured force magnitude was
approximately \(24.5\%\) below the frictionless prediction, with a directional
error of \(14.82^\circ\).

These estimates represent combined transmission losses rather than direct
measurements of individual friction sources. The loss parameters were omitted
from the optimization because they depend on hardware-specific properties
that cannot be identified reliably before fabrication.

Re-scoring the shared design with these efficiencies applied to the
actuator-to-joint torque map, with all poses and design parameters unchanged,
leaves all eight poses force-feasible with a mean coverage loss of \(0.03\).
The loss concentrates in the deeply flexed pose with max loss of \(14\%\)  whose contact forces are
carried mainly by the low-efficiency distal joints, reducing worst-pose coverage
from \(0.88\) to \(0.75\).

\section{Discussion}

The shared tendon routing performs consistently across poses with substantially
different kinematics, increasing worst-pose task-force coverage from \(0.24\)
to \(0.88\) while maintaining the required predicted \(5\,\mathrm{Hz}\)
flexion/extension bandwidth. Although it performs slightly below each
pose-specific design at that design's target pose, it provides more uniform
and higher average performance across the task library.

The optimized architecture assigns the dominant grasping load to one
high-torque tendon and uses three smaller actuators for differential joint
control. This extends the efficiency argument for asymmetric \(N+1\) tendon
arrangements \cite{Pollard2002} by showing that tendon coupling can concentrate
actuation capacity in task-relevant directions without requiring a large
actuator at every joint.

Treating bandwidth as a constraint keeps the force score interpretable while
ensuring minimum dynamic performance. However, the predicted MCP extension
bandwidth of \(5.0\,\mathrm{Hz}\) lies at the imposed limit. Manufacturing
tolerances, friction, tendon compliance, and controller dynamics may reduce
this margin, suggesting that future designs should use a higher threshold.

The experiments confirm the general orientation and scale of the predicted
force set but reveal magnitude and directional errors, particularly for
tendons extending to the DIP joint. This result is consistent with cumulative
friction along longer paths with more routing contacts. Although the
frictionless model is useful for topology and transmission selection, accurate
force prediction requires direction-dependent capstan friction, pretension,
and tendon compliance. Rolling-contact routing elements may further reduce
transmission losses, emphasizing the importance of routing design to overall
hand efficiency.

Comparisons with independently optimized designs also provide a useful
diagnostic. A small performance difference indicates that one shared routing
is sufficient for the task set, whereas a large difference at a particular
pose may motivate changes to the topology, geometric bounds, capstan
dimensions, actuator sizing, or task weighting.

\section{Limitations}

The quasi-static force model assumes rigid links, constant moment arms, and
uniform tendon tension, omitting friction, compliance, hysteresis, backlash,
and pretension. As demonstrated experimentally, these effects can alter both
the magnitude and direction of the realized fingertip force.

The bandwidth model includes actuator torque--speed limits and reflected rotor
inertia but omits controller dynamics, structural flexibility, transmission
friction, and thermal duty-cycle limits. Its predictions are therefore
drivetrain-limited estimates rather than closed-loop bandwidth guarantees.

The task library comprises eight representative poses with prescribed
point-contact force regions rather than forces measured during manipulation.
Force validation was conducted at only one pose and at reduced actuator
limits; broader validation requires experiments across the workspace and under
sustained loading.

Finally, the mixed discrete--continuous search is heuristic and does not
guarantee global optimality. The study also considers one finger in isolation
and therefore does not establish whole-hand grasp stability or force closure.

\section{Conclusion}

This paper presented a task-aware, pose-robust framework that jointly optimizes
tendon routing, moment arms, and capstan dimensions while enforcing exact
joint equilibrium, actuator limits, and drivetrain bandwidth constraints. Its
max--min objective selects one physical design that protects performance at
the weakest required pose.

For a four-DoF finger, the framework produced an asymmetric \(N+1\)
architecture with one high-torque shared-flexion actuator, three smaller
differential actuators, and a separate MCP abduction/adduction actuator.
Across eight poses, this design increased worst-pose task-force coverage from
\(0.24\) to \(0.88\) while satisfying the modelled \(5\,\mathrm{Hz}\)
flexion/extension bandwidth requirement.

Prototype experiments reproduced the general scale and orientation of the
predicted force set but revealed errors caused by unmodeled transmission
losses. Future work will incorporate direction-dependent tendon friction and
compliance, measured task-force data, and whole-hand experimental validation.

\bibliographystyle{IEEEtran}
\bibliography{references}
\end{document}